\documentclass[11pt,a4paper]{article}
\usepackage[hyperref]{naaclhlt2019}
\usepackage{times}
\usepackage{latexsym}
\usepackage{url}

\usepackage{amssymb}
\usepackage{amsmath}
\usepackage{latexsym}
\usepackage{url}
\usepackage{mathtools}
\usepackage{multirow}
\usepackage{xspace}
\usepackage{comment}
\usepackage{flowchart}
\usetikzlibrary{arrows}
\usepackage{booktabs}
\usepackage{subcaption}
\usepackage{hyperref}
\usepackage{enumitem}
\usepackage{multirow}
\usepackage[algo2e,vlined,ruled]{algorithm2e}
\usepackage{tikz}
\usepackage{pgfplots}

\DeclareMathAlphabet{\pazocal}{OMS}{zplm}{m}{n}
\DeclareMathAlphabet{\pazocal}{OMS}{zplm}{m}{n}
\DeclareMathOperator*{\sigm}{sigm}
\DeclareMathOperator*{\argmin}{\arg\!\min}
\DeclareMathOperator*{\argmax}{\arg\!\max}

\aclfinalcopy % Uncomment this line for the final submission
\def\aclpaperid{101} %  Enter the acl Paper ID here

\newcommand{\blue}[1]{\textcolor{blue}{#1}}

\newcommand{\add}[1]{\textcolor{black}{#1}}

\newcommand{\ie}{{\em i.e.,}\xspace}
\newcommand{\eg}{{\em e.g.,}\xspace}

\newcommand{\Ls}{\mathcal{L}}

\newcommand{\Ds}{\pazocal{D}}

\newcommand{\Ni}{({\em i})~}
\newcommand{\Nii}{({\em ii})~}
\newcommand{\Niii}{({\em iii})~}
\newcommand{\Niv}{({\em iv})~}
\newcommand{\Nv}{({\em v})~}
\newcommand{\Na}{({\em a})~}
\newcommand{\Nb}{({\em b})~}
\newcommand{\Nc}{({\em c})~}
\newcommand{\Nd}{({\em d})~}
\newcommand{\Ne}{({\em e})~}
\newcommand{\Nf}{({\em f})~}
\newcommand{\Ng}{({\em g})~}

\newcommand{\sug}{{Suggestion}}

\newcommand{\pol}{{Polite}}

\newcommand{\mrda}{\nobreak{\sc Mrda}}

\title{Adaptation of Hierarchical Structured Models for Speech Act Recognition in Asynchronous Conversation}

\author{Tasnim Mohiuddin$^1$\thanks{All authors contibuted equally.}  \and Thanh-Tung Nguyen$^1$$^\star$  \and Shafiq Joty$^{1,2}$$^\star$ \\
  $^\star$Nanyang Technological University, Singapore \\
$^\dagger$Salesforce Research Asia, Singapore \\
{\tt \{mohi0004@e.,NG0155NG@e.,srjoty@\}ntu.edu.sg}
\\}

\date{}

\begin{document}
\maketitle

\begin{abstract}

We address the problem of speech act recognition (SAR) in asynchronous conversations (forums, emails). Unlike synchronous conversations (e.g., meetings, phone), asynchronous domains lack large labeled datasets to train an effective SAR model. 
In this paper, we propose methods to effectively leverage abundant unlabeled conversational data and the available labeled data from synchronous domains.  
We carry out our research in three main steps. First, we introduce a neural architecture based on hierarchical LSTMs and conditional random fields (CRF) for SAR, and show that our method outperforms existing methods when trained on in-domain data only. Second, we improve our initial SAR models by semi-supervised learning in the form of pretrained word embeddings learned from a large unlabeled conversational corpus. Finally, we employ adversarial training to improve the results further by leveraging the labeled data from synchronous domains and by explicitly modeling the distributional shift in two domains. 

%adversarial domain adaptation methods to  leverage the labeled data from synchronous domains.
%Since asynchronous conversations differ from synchronous ones in their discourse structures, styles, and vocabulary usage, it is not trivial to use the available labeled data from synchronous domains. %, \ie\ conversations where participants interact with each other at different times (\eg\ forums, emails). 

\end{abstract}

\section{Introduction}
%\item \blue{Introduce asynchronous conversation and say how it is different from synchronous ones} 

With the ever-increasing popularity of Internet and mobile technologies, communication media like emails and forums have become an integral part of people's daily life where they discuss events, issues and experiences. Participants interact with each other \emph{asynchronously} in these media by writing at different times, generating a type of conversational discourse that is different from synchronous conversations such as meeting and phone conversations \cite{louis2015conversation}. In the course of the interactions, the participants perform certain communicative acts like asking questions, requesting information, or suggesting something, which are known as \textbf{speech acts} \cite{Austin62}. For example, consider the forum conversation in Figure \ref{fig:example}. The participant who posted the initial comment $\mathbf{C_1}$, describes his situation and asks a couple of \emph{questions}. Other participants \emph{respond} to the initial post with more information and provide \emph{suggestions}. In this process, the participants get into a conversation by taking turns, each of which consists of one or more speech acts. 

%%% Motivating example
\begin{figure}[tb!]
{%\small
\footnotesize
%\begin{description}\setlength\itemsep{5pt}
\begin{description}[style=multiline,leftmargin=0.5cm]
\item[C$_1$:] 
hoping to do the XinJiang Tibet Highway. [\blue{Statement}] \\
Has anyone done it? [\blue{Question}]\\
Am hoping to hire a 4-wheel drive. [\blue{Statement}]\\
I know the roads are bad and I would need an experienced driver and guide - any recommendations? [\blue{Question}]\\
\vspace{-0.7em}
\item[C$_2$:] 
I never done this routine,however i been to Xinjiang twice,in my opinion the local people not friendly, not safe to do this. [\blue{Response}] \\  %  \\
I still have relative stay in Xinjiang, however don't know what they can offer for help... [\blue{Response}]\\
\vspace{-0.7em}
\item[C$_3$:] 
I'm not sure if travelling overland from Xinjiang to Tibet is officially legal yet. [\blue{Response}] \\
You might want to post your question on the North-East Asia branch of Lonely Planet's ThornTree forum for more (useful) answers. [\blue{Suggestion}]\\
\vspace{-0.7em}
\item[C$_4$:] 
a frend and i are trying this route as well, we will likely be in urumuqi and northern part of xinjiang from 8th apr to end apr; looking at doing the xin jiang tibet highway from end apr. (truncated) [\blue{Statement}] \\
contact me at [email] if you want to hook up for possible transport sharing [\blue{Suggestion}]\\
cheers. [\blue{Polite}]\\
%\vspace{-1.5em}
%\item[C$_5$:] 
%April might be a bad time to visit Xinjiang, for weather reasons. [\blue{Suggestion}]\\
%Also note that you'll be running head-on into the Labour Day holiday week in the 1st week of May. [\blue{Statement}] \\
\end{description}
}
\vspace{-1.2em}
\caption{Example of speech acts in a forum thread.}  
\label{fig:example} 
\vspace{-0.5em}
\end{figure}

\textbf{Speech act recognition (SAR)} is an important step towards deep conversational analysis, and can benefit many downstream applications. %including summarization, question answering, and collaborative learning agents \cite{Allen07}.      %\footnote{Taken from \href{https://www.tripadvisor.com.sg/ShowTopic-g294222-i6446-k1073775-XinJiang_Tibet_Highway-Tibet.html}{tripadvisor}} 
%\cite{Joty:2013,louis2015conversation}.
% where information flow is often not sequential as in  or in  (e.g., instant messaging). As a result, discourse structures such as topic structure, coherence structure, and conversational structure in these conversations exhibit different properties from what we observe in monologue or in synchronous conversation 
%\item \blue{Introduce SA recognition and motivate}
%Speech act recognition (SAR) is an important step towards deep conversational analysis \cite{Bangalore06}, and can benefit many  applications including summarization \cite{McKeown07}, question answering \cite{Hong09} and collaborative learning agents \cite{Allen07}
%and artificial companions \cite{Wilks06}. 
 %and artificial companions. 
Availability of large labeled datasets such as the Switchboard-DAMSL (SWBD) \cite{Jurafsky97} and the Meeting Recorder Dialog Act (MRDA) \cite{Dhillon04} corpora has fostered research in data-driven SAR methods in synchronous domains. However, such large corpora are not available in the asynchronous domains, and many of the existing (small-sized) ones use task-specific tagsets as opposed to a standard one. The unavailability of large annotated datasets with standard tagsets is one of the main reasons for SAR not getting much attention in asynchronous domains, and it is often quite expensive to annotate such datasets for each domain of interest.

%bhatia-biyani-mitra:2014:EMNLP2014

SAR methods proposed before the neural `tsunami', \eg\ \cite{qadir-riloff:2011:EMNLP,Minwoo09,tavafi-EtAl:2013:SIGDIAL}, used mostly bag-of-ngram representation (\eg\ unigram, bigram) of a sentence, and most of these methods disregard conversational dependencies (discourse structure) between sentences. Recently, \newcite{joty-hoque:2016}  proposed a neural-CRF framework for SAR in forum conversations. In their approach, a bi-LSTM (trained on the SAR task) first encodes the sentences separately into task-specific embeddings, which are then used in a separate CRF model to capture the conversational dependencies between sentences. They also use labeled data from the MRDA meeting corpus, without which their LSTMs perform worse than simple feed-forward networks. Although their method attempts to model sentence structure (using LSTM) and conversational dependencies (using CRF), the approach has several limitations.    

%,   each sentence  a  The extracted sentence embeddings are then  

%,Joty:Mohiuddin2018}

First, the LSTM-CRF framework was disjoint, and thus cannot be trained end-to-end. Second, when using the MRDA meeting data, their method simply concatenates it with the target domain data assuming they have the same distribution. However,  asynchronous domains (forum, email) differ from synchronous (MRDA) in their underlying conversational structure \cite{louis2015conversation}, in style (spoken vs. written), and in vocabulary usage (meetings on some focused agenda vs. conversations on any topic of interests in a public forum). Therefore, we hypothesize that to make the best use of labeled data from synchronous domains, one needs to model the shift in domains.

In this work, we advance the state-of-the-art of SAR in asynchronous conversations in three main steps. First, we introduce an end-to-end neural architecture based on a hierarchical LSTM encoder with a Softmax or CRF output layer. Second, we improve our initial SAR model by semi-supervised learning in the form of word embeddings learned from a large unlabeled conversational corpus. Most importantly, we adapt our hierarchical LSTM encoder using domain adversarial training \cite{Ganin:2016} to leverage the labeled data from synchronous domains by explicitly modeling the shift in the two domains. 

We evaluate our models on three different asynchronous datasets containing forum and email conversations, and on the MRDA meeting corpus. Our main findings are: \Ni the hierarchical LSTMs outperform existing methods when trained on in-domain data for both synchronous and asynchronous domains, setting a new state-of-the-art; \Nii conversational word embeddings yield significant improvements over off-the-shelf ones; and \Niii domain adversarial training improves the results by inducing domain-invariant features. {The source code, the conversational word embeddings, and the datasets are available at \url{https://ntunlpsg.github.io/demo/project/speech-act/}.}

%at \href{https://ntunlpsg.github.io/demo/project/speech-act/}{https://ntunlpsg.github.io/demo/project/speech-act/}   

%\vspace{-0.2em}
\section{Related Work}
\label{sec:related}
%\vspace{-0.3em}
%\cite{Harshit:2018} should be our base model

%\cite{Chen:2018} most recent paper from SIGIR
Previous studies on \textbf{SAR in asynchronous conversation} have used supervised, semi-supervised and unsupervised methods. \citeauthor{Cohen04} \shortcite{Cohen04} classify emails into acts like `deliver' and `meeting'. Their approach however does not take email context into account. \citeauthor{Carvalho05} \shortcite{Carvalho05} use an iterative algorithm containing two different classifiers: the content classifier that only looks at the content of the message, and the context classifier that takes into account both the content and contextual speech acts in the email thread structure. Other supervised approaches use classifiers and sequence taggers with hand-crafted features \cite{qadir-riloff:2011:EMNLP,tavafi-EtAl:2013:SIGDIAL}.%,Vosoughi-16}. 

\newcite{Minwoo09} use semi-supervised boosting to induce informative patterns from labeled spoken domains (MRDA, SWBD). Given a sentence represented as a set of trees (dependency,  POS tags, n-grams), the boosting algorithm iteratively learns the sub-tree features. %that minimize the errors in the training data.
This approach does not consider the dependencies between the act types, something we successfully exploit in our work. Also, we leverage labeled data from synchronous conversations while adapting our model to account for the domain shift. \newcite{joty-hoque:2016} use a bi-LSTM to encode a sentence, then use a separate CRF to model conversational dependencies. To learn an effective bi-LSTM model, they use the MRDA meeting data; however, without modeling the domain differences. 

The unsupervised methods use variations of Hidden Markov Models (HMM) including HMM-Topic \cite{Alan10}, HMM-Mix \cite{Shafiq11b}, and Mixed Membership \cite{Paul12}.

Several neural methods have been proposed in recent years for \textbf{SAR in synchronous conversations}.  
\citeauthor{Kalchbrenner-2013} \shortcite{Kalchbrenner-2013} use a simple recurrent neural network (RNN) to model sequential dependencies between act types in phone conversations. They use a convolutional network to compose sentence representations from word vectors. \newcite{Lee-NAACL-16} use a similar model,  but also experiment with RNNs to compose sentence representations. \newcite{Khanpour-coling-16} use a stacked LSTM to compose word vectors into a sentence vector. \newcite{Harshit:2018} also use a hierarchical LSTM-CRF. %\newcite{Chen:2018} propose a similar framework, but apply attention to get the sentence representations. 
However, none of these methods were applied to asynchronous conversations, where not much labeled data is available. 
Also to the best of our knowledge, no prior work attempted to do domain adaptation from the synchronous conversation, which is our main contribution in this paper.

\section{The Base Model}
\label{sec:model}
%\vspace{-0.3em}
%In this section, we present our base models for SAR in asynchronous conversation. In Section \ref{sec:adap}, we present our methods to adapt this model.  

\vspace{-0.2em}
%\subsection{The Base Model}

We use a bidirectional long short-term memory or bi-LSTM  \cite{Hochreiter:1997} to encode each sentence into a vector representation. Given an input sentence $\mathbf{x}_i = (w_1, \cdots, w_m)$ of length $m$, we first map each word $w_t$ to its corresponding vector representation $\mathbf{v}_t$ by looking up the word embedding matrix. The LSTM recurrent layer then computes a compositional representation ${\mathbf{z}}_t$ at every time step $t$ by performing nonlinear transformations of the current input $\mathbf{v}_t$ and the output of the previous time step ${\mathbf{z}}_{t-1}$. The output of the last time step ${\mathbf{z}_m}$ is considered as the representation of the sentence. A bi-LSTM composes a sentence in two directions:  left-to-right and  right-to-left, yielding a representation $\mathbf{h}_i = [\overrightarrow{\mathbf{z}_m}; \overleftarrow{\mathbf{z}_m}]$, where `;' denotes concatenation. Similar to \cite{joty-hoque:2016}, we could use $\mathbf{h}_i$ to classify sentence $\mathbf{x}_i$ into one of the speech act types using a Softmax output layer. However, in that case, we would disregard the discourse-level dependencies between sentences in a conversation. To take conversational dependencies into account, we explore two methods as we describe below.

\subsection{Hierarchical LSTM} 

\begin{figure}[t!]
  \centering
%\scalebox{0.95}
{
  \includegraphics[width=1\linewidth]{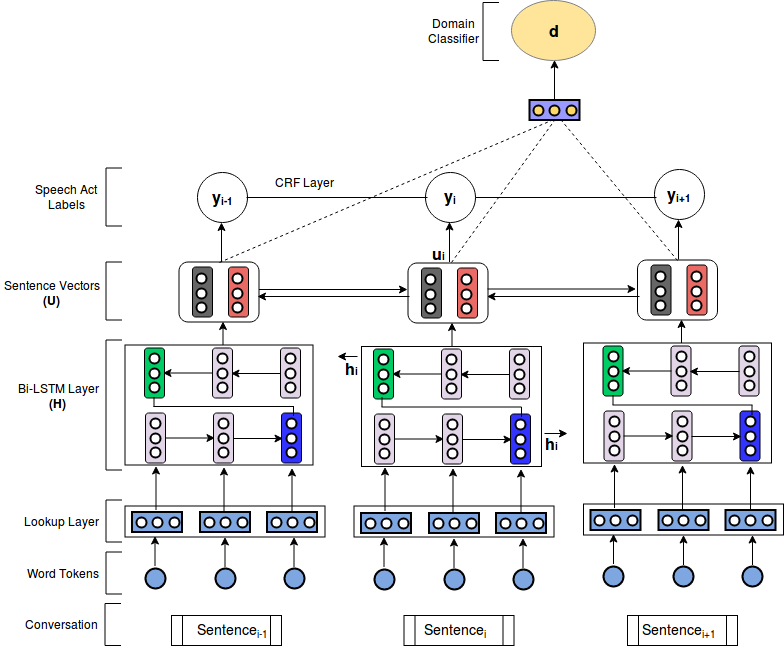}}
  %\vspace{-0.1em}
\caption{Hierarchical bi-LSTM-CRF model with domain adversarial training for speech act recognition.}
\label{fig:speech-act-model}
%\vspace{-0.5em}
\end{figure}

We consider a conversation as a sequence of utterances (sentences). Given an input sequence of $n$ sentences $\mathbf{X} = (\mathbf{x}_1, \cdots, \mathbf{x}_n)$, the sentence-level bi-LSTM generates a sequence of $n$ vectors $\mathbf{H} = (\mathbf{h}_1, \cdots, \mathbf{h}_n)$. To consider interdependencies between sentences, we place another bi-LSTM layer on top of $\mathbf{H}$ to connect the sentence vectors sequentially in both directions, and encode each sentence within its left and right contexts. As shown in Figure \ref{fig:speech-act-model}, the upper bi-LSTM combines the current input $\mathbf{h}_i$ with its previous hidden state $\overrightarrow{\mathbf{u}}_{i-1}$ (resp., $\overleftarrow{\mathbf{u}}_{i+1}$) to generate a representation for the current sentence $\overrightarrow{\mathbf{u}}_{i}$ (resp., $\overleftarrow{\mathbf{u}}_{i}$). The hierarchically encoded sentence vectors $\mathbf{U} = (\mathbf{u}_1, \cdots, \mathbf{u}_n)$ (where $\mathbf{u}_i = [\overrightarrow{\mathbf{u}_i}; \overleftarrow{\mathbf{u}_i}]$) are fed into a Softmax classifier for speech act classification.

%\vspace{-0.5em}

%\footnote{In theory,  a right-to-left representation should add useful information. However, given the small labeled datasets, we did not observe any clear gain for a bidirectional LSTM at the upper level.}

\vspace{-0.5em}
%\small
\begin{equation*}
p(y_i = k|\mathbf{X}, W, \theta) = \frac{\exp~(\mathbf{w}_k^T\mathbf{u}_i)} {\sum_{k=1}^{K} \exp~({\mathbf{w}_k^T\mathbf{u}_i)}} \label{softmax}
\end{equation*}
\normalsize

\noindent  where $W$ are the classifier weights, and $\theta$ are the parameters of the hierarchical LSTM encoder. We train the model by minimizing the cross entropy:

%\small
\vspace{-0.5em}
\begin{eqnarray*}
\Ls_c(W, \theta) = - \sum_{i=1}^{n} \sum_{k=1}^{K} y_{i,k} \log p(y_i = k|\mathbf{X}, W, \theta) \label{logloss}
\end{eqnarray*}
\vspace{-0.5em}
\normalsize

\noindent with $y_{i,k}$ being the one-hot encoding of the label.  

\subsection{Hierarchical LSTM with CRF}

The hierarchical LSTM (H-LSTM) captures contextual information by propagating information through hidden layers, and has been shown to be effective in similar tasks such as context encoding in dialog systems \cite{Serban:2016}. Despite this, its modeling strength is limited compared to structured models that use global inference to model consistency in the output, especially when there are strong dependencies between output labels \cite{Collobert11}. Therefore, instead of classifying sentences independently with a Softmax layer, our second method is to model them jointly with a CRF layer \cite{Lafferty01}. For an input-output sequence pair $(\mathbf{X}, \mathbf{y})$, we define the joint probability distribution: 

\small
\vspace{-0.5em}
\begin{eqnarray}
p(\mathbf{y}|\mathbf{X}) = \frac{1}{Z(\mathbf{U},A,V,\theta)} \prod_{i=1}^n \underbrace{\psi_n(y_i|\mathbf{u}_i,V)}_\text{node factor} \hspace{-0.05cm} \prod_{i=0}^n \underbrace{\psi_e(y_{i,i+1}|{A})}_\text{edge factor} \label{eq:pcrf} \nonumber
\end{eqnarray}
\vspace{-0.5em}
\normalsize

\noindent where $\mathbf{U} = (\mathbf{u}_1, \cdots, \mathbf{u}_n)$ is the hierarchically encoded sentence vectors as before, and $\psi_n(y_i=k|\mathbf{u}_i, V) = \exp (V_{k}^T\mathbf{u}_i)$ is the node-level score with $V$ being the weight matrix, $\psi_e$ is the transition matrix parameterized by $A$, and $Z(.)$ is the global normalization constant that ensures a valid probability distribution. The cross entropy loss for the $(\mathbf{X}, \mathbf{y})$ sequence pair can be written as:

\small
\vspace{-0.5em}
\begin{eqnarray}
\Ls_c(V,A,\theta) \hspace{-0.05cm}= \hspace{-0.05cm}- \hspace{-0.05cm} \sum_{i=1}^n \hspace{-0.05cm}\log \psi_n(y_i|\mathbf{u}_i,V)  - \hspace{-0.05cm} \hspace{-0.05cm}\sum_{i=0}^n \log A_{i,i+1} 
  \hspace{-0.05cm}+\hspace{-0.05cm}\log Z \nonumber  
\end{eqnarray}
\vspace{-0.5em}
\normalsize

\noindent We use Viterbi decoding to infer the most probable tag sequence for an input sequence of sentences, $\mathbf{y}*$ $=$ $\argmax_{\mathbf{y}} p(\mathbf{y}|\mathbf{X},V,A,\theta)$. {We will demonstrate later in our experiments that a CRF layer helps the H-LSTM to adapt quickly (\ie\ with less labeled data) to a target domain by exploiting the tag dependencies in the source domain.}

%\noindent where $\psi_n(y_i=k|\mathbf{u}_i, V) = \exp (V_{k}^T\mathbf{u}_i)$ is the node-level score with $V$ being the weight matrix, $\psi_e$ is the transition matrix parameterized by $A$, and $Z(.)$ is the global normalization constant that ensures a valid probability distribution. 

%We explore two possible ways to build a sentence-level SAR model from these sentence representations. Our first approach is similar to \cite{joty-hoque:2016}, where we classify each sentence independently. However, as remarked earlier, this approach does not consider the discourse structure or conversational dependencies between the sentences in a conversation.  

%\vspace{-0.3em}
\section{Adaptation Methods} 
\label{sec:adap}
%\vspace{-0.3em}
The hierarchical models have many parameters. Given enough training data, they should be able to encode a sentence, capturing its syntactic and semantic properties, and discourse-level dependencies. However, when it comes to SAR in asynchronous domains, not many large annotated corpora are available. Because of the large number of parameters, the models usually overfit when trained on small datasets of asynchronous conversations (shown in Sec. \ref{sec:exper}). We propose two solutions to address this problem. Our first (simple but effective) solution  is to leverage large unlabeled conversational corpus to learn better task-agnostic word embeddings, and use it to initialize our models for better generalization. In the interests of coherence, we present this method in Section \ref{sec:corpora}. 

%        
%Our first solution (that works surprisingly well) 

Our second solution is to leverage data from synchronous domains  for which large annotated corpus is available (\eg\ MRDA corpus). However, as we will see, simple concatenation of the datasets is not quite effective in our case, because the conversations in synchronous and asynchronous domains differ in their conversational structures, modality (spoken vs. written), and vocabulary usage. To get the best out of the available synchronous domain data, we need to adapt our  models by explicitly modeling the domain shift. More precisely, our goal is to adapt the hierarchical encoder so that it learns to encode sentence representations $\mathbf{U}$ (\ie\ features used for classification) that is not only discriminative for the act classification, but also invariant across the domains. We propose to use the \textbf{domain adversarial training} proposed by \newcite{Ganin:2016}.

Let $\Ds_S = \{\mathbf{X}_p, \mathbf{y}_{p} \}_{p=1}^{P}$ denote the set of $P$ labeled training conversations  in the \textbf{source} domain (MRDA). We consider two  adaptation scenarios. 

%\begin{itemize}[leftmargin=*]

\textbf{\Ni Unsupervised adaptation:} In this scenario, we have only \emph{unlabeled} examples in the \textbf{target} domain (\eg\ forum). Let $\Ds_T^u = \{\mathbf{X}_p\}_{p=P+1}^{Q}$ be the set of $(Q - P - 1)$ unlabeled training instances in the target domain with $Q$ being the total number of training instances in the two domains.   

\textbf{\Nii Semi-supervised/supervised adaptation:} In addition to the unlabeled instances $\Ds_T^u$, here we have access to some \emph{labeled} training instances in the target domain,  $\Ds_{T}^l = \{\mathbf{X}_p, \mathbf{y}_p\}_{p=Q+1}^{R}$, with $R$ being the total number of training examples in the two domains. Depending on the amount of labeled data in the target domain, this setting is referred to as semi-supervised or supervised adaptation. 

%\end{itemize}

\subsection{Unsupervised Adaptation} \label{subsec:unsup-adap} 
The dashed lines in Figure \ref{fig:speech-act-model} show the extension of our base model for adaptation. The input conversation $\mathbf{X}$ is sampled either from a synchronous domain (\eg\ meeting) or from an asynchronous  domain (\eg\ forum). Our goal is to adapt the H-LSTM encoder (parameterized by $\theta$) to generate $\mathbf{U}$ such that it is not only informative for the SAR task but also invariant across domains. Upon achieving this, we can use the adapted encoder to encode a target sentence, and use the source classifier (Softmax or CRF) to classify the sentences.

%As before, we pass $\mathbf{X}$ through the H-LSTM encoder (parameterized by $\theta$) to get $\mathbf{U}$. 
%Our goal is to adapt the encoder  

We achieve this by adding a domain discriminator (dashed lines in Figure \ref{fig:speech-act-model}), another neural network that takes $\mathbf{U}$ as input, and tries to discriminate the domains of the input conversation $\mathbf{X}$ (\eg\ meeting vs. forum). The output of the  discriminator is defined by a sigmoid function: 

%\small
\vspace{-0.5em}
\begin{equation}
\hat{d}_{\omega} = p(d = 1|\mathbf{u}_i, \omega, \theta) = \sigm (\mathbf{w}_d^T \mathbf{h}_d)
\end{equation}
\vspace{-0.5em}
\normalsize

\noindent where $d\in\{0,1\}$ denotes the domain ($1$ for meeting, $0$ for forum), $\mathbf{w}_d$ are the final layer weights of the discriminator, and $\mathbf{h}_d = g(U_d \mathbf{u}_i)$ defines the hidden layer of the discriminator with $U_d$ being the layer weights, and $g(.)$ being the activations. We use cross entropy as the discrimination loss: 

%For domain adaptation,  $\mathbf{z}$  \add{}.

%\small
\vspace{-0.5em}
\begin{eqnarray}
\Ls_d(\omega, \theta) =  - d \log \hat{d}_{\omega} - (1-d) \log \left(1- \hat{d}_{\omega} \right) \label{eq:dis-loss}
\end{eqnarray}
\vspace{-0.5em}
\normalsize

\noindent The composite network has three players: the hierarchical \textbf{LSTM encoder},  the \textbf{classifier} (Softmax or CRF), and the domain \textbf{discriminator}. During training, the encoder and the classifier play a co-operative game, while the encoder and the discriminator play an adversarial game.  The training objective ${\Ls(W, \theta, \omega)}$ of the composite model  is:

%We can write the overall training objective as 

\small
\vspace{-0.5em}
\begin{eqnarray}
%\Ls(W, \theta, \omega) =& \nonumber \\ %\vspace{-1em}
&\hspace{-0.5cm}  \underbrace{\sum_{p=1}^P \Ls_c^p (W, \theta)}_\text{act classif (src)}   
- \lambda \Big[ \underbrace{\sum_{p=1}^P \Ls_d^p(\omega, \theta)}_\text{domain disc (src)} + \underbrace{\sum_{p=P+1}^Q \Ls_d^p(\omega, \theta)}_\text{domain disc (tar)} \Big]  \label{loss}
\end{eqnarray}
\vspace{-0.5em}
\normalsize

\noindent where $\theta$ are the parameters of the encoder, $W$ are the classifier weights, and $\omega = \{U_d,\mathbf{w}_d\}$ are the parameters of the discriminator.\footnote{For  simplicity, we describe adaptation of the encoder with Softmax output, but this generalizes naturally to CRF.} The hyper-parameter $\lambda$ controls the relative strength of the act classifier and the discriminator. We learn $\theta$ that optimizes the following \textbf{min-max} criterion: 

%In training, we look for parameter values that satisfy the following \emph{min-max} optimization criterion:

\vspace{-0.5em}
\begin{eqnarray}
\theta^* = \argmin_{W, \theta} \max_{U_d,\mathbf{w}_d} \Ls(W, \theta,\omega) 
\end{eqnarray}
\vspace{-0.5em}
\normalsize

%which involves a maximization (gradient ascent) with respect to $\{U_d,\mathbf{w}_d\}$ and a minimization (gradient descent) with respect to $\theta$ and $W$. 

%Maximizing $\Ls(W, \theta,\omega)$ with respect to $\{U_d,\mathbf{w}_d\}$ is equivalent to minimizing the discriminator loss $\Ls_d(\omega, \theta)$ in Eq. \ref{eq:dis-loss}. 

Note that the updates of the shared encoder for the two networks (classifier and discriminator)  work \emph{adversarially} with respect to each other. Algorithm \ref{alg:training} provides pseudocode of our training method. The main challenge in adversarial training is to balance the networks \cite{ArjovskyCB17}. In our experiments, we found the discriminator to be weaker initially. %If one player becomes smarter, its loss to the shared layer becomes useless, and the loss of the weaker one overwhelms the other, causing training to fail . 
 To balance the two components, we would need the error signals from the discriminator to be fairly weak initially, with full power unleashed only as the classification errors start to dominate. We follow the weighting schedule proposed in \cite[p. 21]{Ganin:2016}, which initializes $\lambda$ to $0$, and then changes it gradually to $1$ as training progresses. %I.e., we start training the task classifier first, and we gradually add the discriminator's loss. 

\begin{algorithm2e}[h!]
\small
\SetKwInOut{Input}{Input}\SetKwInOut{Output}{Output}
%\SetAlgoLined
\SetAlgoNoLine
\SetNlSkip{0em}
\Input{Data $\Ds_S$, $\Ds_T^u$, and batch size $b$}
\Output{Adapted model parameters $\theta$, W}
1. Initialize model parameters; \\
2. \Repeat {convergence}{ 
        
		\hspace{-0.4em}\Na Randomly sample $\frac{b}{2}$ labeled examples from $\Ds_S$ \\
		\hspace{-0.4em}\Nb Randomly sample $\frac{b}{2}$ unlabeled examples from $\Ds_T^u$ \\
		\hspace{-0.4em}\Nc Compute $\Ls_c(W, \theta)$ and $\Ls_d(\omega, \theta)$ \\
        \hspace{-0.4em}\Nd \add{Set $\lambda = \frac{2}{1 + \exp(-10*p)} - 1$; $p$ is the training progress linearly changing form $0$ to $1$.}\\
        \tcp{Classifier \& Encoder}
		\hspace{-0.4em}\Ne Take a gradient step for $\frac{2}{b} \nabla_{W,\theta} \Ls_c(W, \theta) $   \\
        \tcp{Discriminator}
		\hspace{-0.4em}\Nf Take a gradient step for $\frac{2\lambda}{b} \nabla_{U_d,\mathbf{w}_d} \Ls_d(\omega, \theta)$\\
%        where $\lambda$ is the domain adaptation parameter and computed using the equation: $\lambda = \frac{2}{1 + \exp(-10*p)} - 1$, here $p$ is the training progress linearly changing form $0$ to $1$.\\
        \tcp{Gradient reversal}
		\hspace{-0.4em}\Ng Take a gradient step for $- \frac{2\lambda}{b} \nabla_{\theta} \Ls_d(\omega, \theta)$ 
   }
\caption{Adversarial training with SGD.}
\label{alg:training}
\end{algorithm2e}

\subsection{Semi-supervised/supervised Adaptation}

It is straight-forward to extend our adaptation method to a semi-supervised/supervised setting. Similar to the instances in the source domain, the labeled instances in the target domain $\Ds_{T}^l$ are used for act classification and domain discrimination. The total training loss $\Ls(W, \theta, \omega)$ in this case is 

% each minibatch during training is formed by labeled instances from both $\Ds_S$ and $\Ds_{T^*}$, and unlabeled instances from $\Ds_T$.

\small
\vspace{-0.5em}
\begin{eqnarray}
\underbrace{\sum_{p=1}^P \Ls_c^p (W, \theta)}_\text{act classif. (source)}  + \hspace{-0.7em} \underbrace{\sum_{p=Q+1}^R \hspace{-0.9em} \Ls_c^p (W, \theta)}_\text{\textit{\blue{act classif. (target)}}} - 
\lambda \Big[\hspace{-0.4em}\underbrace{\sum_{p=1}^P \Ls_d^p(\omega, \theta)}_\text{dom classif. (source)} \hspace{-0.4em} + \hspace{-0.8em}\underbrace{\sum_{n=P+1}^{R} \hspace{-0.6em} \Ls_d^p(\omega, \theta)}_\text{\blue{dom classif. (target)}} \hspace{-0.3em}\Big] \nonumber \label{loss}
\end{eqnarray}
\vspace{-0.5em}
\normalsize

\noindent \add{where the second term is the classification loss on the target dataset $\Ds_{T}^l$, and the last term is the discrimination loss on both labeled and unlabeled data in the target domain.} %We modify the training algorithm accordingly. Specifically, each minibatch in SGD training is formed by labeled instances from both $\Ds_S$ and $\Ds_{T}^l$, and unlabeled instances from $\Ds_T^u$.  

%\section{Our Approach} 

%\vspace{-0.3em}
\section{Corpora}
\label{sec:corpora}
%\vspace{-0.3em}
We now describe the datasets and the act tagset that we use, and the conversational word embeddings that we learn from a large unlabeled corpus.  

\subsection{Labeled Datasets} 

As mentioned, asynchronous domains lack large corpora that are annotated with a standard speech act tagset. \newcite{Minwoo09} annotated sentences in TripAdvisor \textbf{(TA) forum} threads with the standard 12 act types defined in MRDA. They also remapped the \textbf{BC3 email} corpus \cite{JanAAAI08} according to these tags. Subsequent studies \cite{tavafi-EtAl:2013:SIGDIAL,oya-carenini:2014:W14-43,joty-hoque:2016} used these datasets but grouped the 12 acts into 5 coarser classes. \newcite{joty-hoque:2016} also created a new dataset of QatarLiving\footnote{\href{http://www.qatarliving.com/}{http://www.qatarliving.com/}} \textbf{forum} threads called \textbf{QC3}.\footnote{\href{https://ntunlpsg.github.io/project/speech-act/}{https://ntunlpsg.github.io/project/speech-act/}} We use these three asynchronous datasets in our experiments. {For our experiments on synchronous domains, we use the \textbf{MRDA meeting} corpus that was also used in  related studies \cite{Minwoo09,joty-hoque:2016}. Tables \ref{tab:corpora} and \ref{tab:tag_distribution} show some basic statistics of the datasets and the tag distributions. {Note that the tagset used by us and other related studies in asynchronous (written) conversation is different from the one used in synchronous spoken conversations \cite{Lee-NAACL-16,Khanpour-coling-16,Harshit:2018}. The later tagset contains acts like \emph{backchannel}, \emph{filter} and \emph{disruption} that are more specific to speech.}} 

%The later cluster of works uses a tagset that contains statement, question, backchannel, filler, and  acts, and is more specific to speech.}

\begin{table}[t!]
\begin{center}
\scalebox{0.72}{
\begin{tabular}{l|lll|l}
\toprule
 & \multicolumn{3}{c|}{\textbf{Asynchronous}} &  {\textbf{Synchronous}}\\
 \hline
 & {TA} & {BC3} & {\add{QC3}} & \mrda \\
\hline
Total \# of conversations  & 200 & 39 & \add{47} & 73 \\
Avg. \# of comments/conv & 4.02 & 6.54 & \add{13.32} & N.A \\
Avg. \# of sentences/conv & 18.56 & 34.15 & \add{33.28} & 955.10\\
Avg. \# of words/sen & 14.90 & 12.61 & \add{19.78} & 10.11 \\
\bottomrule
\end{tabular}}
\vspace{-0.3em}
\caption{Basic statistics about our corpora.}
\label{tab:corpora}
\end{center}
%\vspace{-0.5em}
\end{table}

\begin{table}[tb!]
\begin{center}
\scalebox{0.8}{
\begin{tabular}{l|l|lll|l}
\toprule
 & & \multicolumn{3}{c|}{\textbf{Asynchronous}} & {\textbf{Synchronous}}\\
 \hline
 {Tag} & {Description} & {TA} & {BC3} & {\add{QC3}} & {\mrda} \\
\hline
SU & Suggestion & 7.71 & 5.48 & 17.38  & 5.97\\ 
R & Response & 2.4 & 3.75 & 5.24 & 15.63\\
Q & Questions & 14.71 & 8.41 & 12.59 & 8.62\\
P & Polite  & 9.57 & 8.63 & 6.13 & 3.77\\
ST & Statement & 65.62 & 73.72 & 58.66  & 66.00\\
\bottomrule
\end{tabular}}
\vspace{-0.3em}
\caption{Distribution of speech acts in our corpora.}
\label{tab:tag_distribution}
\end{center}
%\vspace{-1em}
\end{table}

{The train-dev-test splits of the asynchronous datasets are done uniformly at random at the \textbf{conversation level}. Since the asynchronous datasets are quite small in size, to have a reliable test set, we create the train:test splits with an equal number of conversations (Table \ref{tab:con_dataset}). \newcite{joty-hoque:2016} also created conversation level datasets to train and test their CRF models. Their test sets however contain only 20\% of the conversations, providing only 5 conversations for QC3 and BC3, and 20 for TA. Our experiments on these small test sets showed unstable results for all the models. Therefore, we use a larger test set (50\%), and we report more general  results on the whole corpus based on \textbf{2-fold} cross-validation, where the second fold was created by interchanging the train and test splits in Table \ref{tab:con_dataset}.
%\footnote{Less training data in the target domain also simulates a more realistic scenario where domain adaptation is needed} 
The same development set was used to tune the hyperparameters of the models for experiments on each fold. For experiments on MRDA, we use the same train:test:dev split as in \cite{Minwoo09,joty-hoque:2016}.}

%This is in contrast to \newcite{joty-hoque:2016} that used 20\% for testing. 
%  In our initial experiments, we used the same split as in \cite{joty-hoque:2016}. Although our models performed well, the results were unstable on those small test sets. Therefore,   

\begin{table}[t!]
\centering
\scalebox{0.85}{\begin{tabular}{l|l|l|l|l}
\toprule
 &  &  \textbf{Train} & \textbf{Dev.} & \textbf{Test} \\
\hline
\textbf{Source} & \mrda\ & 59 (50865) & 6 (8366) & 8 (10492) \\
\hline
\multirow{3}{*}{\textbf{Target}}  & TA   & 90 (1667)    & 20 (427) & 90 (1617) \\
& QC3  & 20 (675)    & 7  (230) & 20 (660)  \\
& {BC3}  & {16 (557)}    & {7 (233)} & {16 (542)}  \\
\hline
\add{Total} & & 126 (2899) & 34 (890) & 126 (2819) \\
\bottomrule
\end{tabular}}
\vspace{-0.3em}
\caption{\label{tab:con_dataset} Train, dev. and test sets for the datasets. Numbers in parentheses indicate the number of sentences.}
%Datasets to train, validate and test our models. 
%The splits are done uniformly at random at the conversation level. 
%\vspace{-1.2em}
\end{table}

\subsection{Conversational Word Embeddings} \label{subsec:word-emb}
%\vspace{-0.5em}

One simple and effective approach to semi-supervised learning is to use word embeddings pretrained from a large unlabeled corpus. In our work, we use {generic} \emph{off-the-shelf} pretrained embeddings  to boost the performance of our models. In addition, we have also trained word embeddings from a large conversational corpus to get more relevant \textbf{conversational} word embeddings. %Later in our experiments we will show that the \emph{conversational} word embeddings are more effective than the generic ones since they are trained on similar datasets.
%Many recent studies have shown that the pretrained embeddings improve the performance on supervised tasks \cite{schnabel-EtAl:2015:EMNLP}. 

We use Glove \cite{pennington-socher-manning:2014:EMNLP2014} to train our word embeddings from a corpus that contains 24K {email} threads from W3C (w3c.org), 25K threads from TripAdvisor,  220K threads from QatarLiving, and all conversations from SWBD and MRDA  (a total of 120M tokens). Table \ref{tab:conv_embedding} shows some statistics of the datasets used for training the conversational word embeddings. {We also trained skip-gram word2vec \cite{Mikolov:2013}, but its performance was worse than Glove.}  

\begin{table}[t!]
\begin{center}
\scalebox{0.83}{
\begin{tabular}{l|ccc}
\toprule
& \# of Threads & \# of Tokens & \# of Words \\
% & \multicolumn{3}{c|}{\textbf{Asynchronous}} &  \multicolumn{2}{c}{\textbf{Synchronous}}\\
 \hline
{W3C}  & 23,940 & 21,465,830 &  546,921\\
{TripAdvisor} & 25,000  & 2,037,239 & 127,333   \\
{\add{Qatar Living}} & 219,690 & 103,255,922  & 1,157,757 \\
\midrule
MRDA & - & 675,110 & 18,514 \\
SWBD & - & 1,131,516  & 57,075 \\
\bottomrule
\end{tabular}}
\vspace{-0.3em}
\caption{\label{tab:conv_embedding} \add{Datasets and their statistics used for training the conversational word embeddings.}
}
\end{center}
%\vspace{-0.7em}
\end{table}

\

%\vspace{-1em}
\section{Experiments}
\label{sec:exper}
%\vspace{-0.3em}
%In this section, we present our experiments. 

\begin{table*}[h!]
\centering
%\small
\vspace{-0.5em}
\scalebox{0.90}{
%\begin{tabular}{l@{\hspace{0.1cm}}|c@{\hspace{0.1cm}}|c@{\hspace{0.1cm}}|c@{\hspace{0.1cm}}|c}
\begin{tabular}{l|c|c|c|c}
\toprule
& \textbf{QC3}  & \textbf{TA} & \textbf{BC3} & \textbf{MRDA} \\
\hline
\add{SVM$_\text{c-gl}$}  & $16.96_{\pm0.00}$  & $20.17_{\pm0.00}$  & $17.20_{\pm0.00}$ & $31.47_{\pm0.00}$ \\
\add{FFN$_{\text{c-gl}}$} & $48.29_{\pm0.25}$  & $61.36_{\pm0.21}$  & $39.58_{\pm0.26}$ & $71.12_{\pm0.13}$ \\
{FFN$_{\text{skip-th}}$} & $50.80_{\pm1.21}$  & $61.44_{\pm0.92}$  & $47.67_{\pm0.74}$ &  $71.73_{\pm0.48}$\\
\midrule
{B-LSTM$_\text{rand}$} & $50.25_{\pm0.57}$  & $62.11_{\pm0.64}$  & $45.08_{\pm1.03}$ & $70.72_{\pm0.02}$	\\
\add{B-LSTM$_\text{gl}$} & $53.21_{\pm0.77}$  & $63.23_{\pm0.80}$  & $49.04_{\pm0.90}$ & $72.23_{\pm0.18}$\\
\add{B-GRU$_\text{c-gl}$} & $60.50_{\pm0.36}$  & $67.23_{\pm0.76}$  & $55.45_{\pm1.05}$ & $72.04_{\pm0.35}$\\
\add{B-LSTM$_\text{c-gl}$} & $\mathbf{61.01_ {\pm 0.60}}$ & $67.23_{\pm0.70}$ & $55.32_{\pm0.68}$ & $72.42_{\pm0.14}$\\
\add{S-LSTM$_\text{c-gl}$} & $56.70_{\pm0.58}$ & $62.28_{\pm1.23}$ & $52.31_{\pm0.86}$ & $71.32_{\pm0.28}$\\
\midrule
H-LSTM$_\text{c-gl}$ & $60.76_ {\pm 0.99}$ & $\mathbf{68.38_{\pm0.65}}$ & $\mathbf{57.17_{\pm 0.87}}$ & $\mathbf{72.91_{\pm0.14}}$\\
H-LSTM-CRF$_\text{c-gl}$ & $59.83_ {\pm 1.27}$ & $68.10_{\pm0.68}$ & $56.37_{\pm 0.61}$& $72.77_{\pm0.17}$ \\
\bottomrule
\end{tabular}
}
\vspace{-0.5em}
\caption{Macro-$F_1$ scores for \textbf{in-domain training}.} 
\label{tab:indom-results} 
\end{table*}

{We followed similar preprocessing steps as \newcite{joty-hoque:2016}; specifically: normalize all characters to lower case, spell out digits and URLs, and tokenize the texts using TweetNLP \cite{gimpel2011part}. For performance comparison, we use  \textbf{accuracy} and \textbf{macro-}$\mathbf{F_1}$. Like other related studies, we consider $\text{macro-F}_1$ as the main metric (more appropriate when class distributions are imbalanced), and select our model based on the best ${F}_1$ on the development set. Due to space limitations, we report only  $\text{macro-F}_1$ here. Please refer to the Appendix for the accuracy numbers.}

\subsection{Experiments on In-domain Training} \label{sec:eval-in-domain}

We first evaluate our base models on in-domain datasets by comparing with state-of-the-art models. In the next subsection, we evaluate our adaptation method in the three adaptation scenarios.

\vspace{-0.3em}
\paragraph{Settings.}

%Our main objective is to evaluate the proposed speech act recognizer on asynchronous conversations. For this, we evaluate our models on the forum and email datasets introduced earlier. In addition, we validate our sentence encoding approach on the \mrda\ meeting corpus.

To validate the efficacy of our model, we compare it with two baselines: a Support Vector Machine (\textbf{SVM}) and a  feed-forward network (\textbf{FFN}). {In one setting, we use the concatenated word vectors as the input sentence representation, while in another, we use the pretrained skip-thought vectors \cite{Kiros:2015}.}  We also compare our models with the {bi-LSTM} (\textbf{B-LSTM}) model of \newcite{joty-hoque:2016} and the stacked LSTM (\textbf{S-LSTM}) of \newcite{Khanpour-coling-16}.

%Multi-layer Perceptron (\textbf{MLP}) with one hidden layer. 

We use the Adam optimizer \cite{KingmaB14} with a learning rate of 0.001, and use dropout  to avoid over-fitting. We use the Xavier initializer \cite{GlorotAISTATS2010} to initialize the weights, and uniform $\mathcal{U}(-0.05, 0.05)$ to initialize the word vectors randomly. For pretrained word embeddings, we experiment with off-the-shelf embeddings that come with Glove as well as with our conversational word embeddings. For both random and pretrained initialization, we \textbf{fine-tune} our word embeddings on the SAR task.

We construct sequences from the \textbf{chronological order} of the sentences in a conversation. Since MRDA conversations are much longer compared to those in asynchronous domains (955 vs. 18-34 sentences in Table \ref{tab:corpora}), we split the MRDA  conversations into smaller parts containing  a maximum of 100 sentences.\footnote{{In a different setting, we created sequences by connecting each non-initial comment with the initial comment generating many 2-comment sequences. This is considering the fact that in many QA forums, users mostly answer to the questions asked in the initial post. In our experiments on in-domain training, we found this competitive with our `one long-chain' structure. However, the adaptation  in this setting was much worse because of the mismatch in discourse structures of synchronous and asynchronous conversations.}} The number of epochs and batch size were fixed to 30 and 5 (conversations), respectively. We ran each experiment $5$ times, each time with a different random seed, and report the average of the (2-fold$\times$5=10) runs along with the standard deviation. Recently, \newcite{Crane-2018} show that the main source of variability in results for neural models come from the random seed, and the author has recommended to report the distribution of results from a range of seeds.

%\red{Say what makes a sequence in \mrda}For MRDA, we create conversational data by splitting each speech dialogue into   with maximum size 100 sentences each.

%H-LSTM outperforms other models in every dataset except QC3 where the difference is very small.\footnote{There is no significant difference between H-LSTM and B-LSTM in accuracy also as can be seen in the Appendix.} 

%{We also notice that the H-LSTM-CRF outperforms other models except H-LSTM. The reason could be that the contextual dependency is already captured by the LSTM layers and the data may be a bit small for CRF to capture anything more.}    

%For both of the models, we use the pretrained \emph{conversational} glove embeddings for initialization.

\vspace{-0.5em}
\paragraph{Results.}

%For performance comparison, we use both \textbf{accuracy} and \textbf{macro-averaged} $\mathbf{F_1}$ score. Accuracy gives the overall performance of a classifier but could be biased to most populated ones. From Table \ref{tab:tag_distribution}, we see that the distribution of speech acts in our corpora is not balanced. On the other hand, macro-averaged $F_1$ weights every class equally, and is not influenced by class imbalance. For space limitations, we only show $F_1$ scores here and report both accuracy and $F_1$ score in the supplementary document.

We present the results in Table \ref{tab:indom-results}. From the \textbf{first block} of results, we notice that both SVM and FFN baselines perform poorly compared to other models that tune the word embeddings and learn the sentence representation on the SAR task. 

The \textbf{second block} contains five LSTM variants: \Ni \textbf{B-LSTM$_\text{rand}$}, referring to bi-LSTM with \emph{random} initialization;  \Nii \textbf{B-LSTM$_\text{gl}$}, referring to bi-LSTM initialized with \emph{off-the-shelf} Glove embeddings; \Niii \textbf{B-GRU$_\text{c-gl}$}, referring to bidirectional Gated Recurrent Unit \cite{cho-al-emnlp14} initialized with our \emph{conversational} Glove; \Niv \textbf{B-LSTM$_\text{c-gl}$}, referring to bi-LSTM initialized with \emph{conversational} Glove, and \Nv \textbf{S-LSTM$_\text{c-gl}$}, referring to a 2-layer stacked LSTM with \emph{conversational} Glove.\footnote{Increasing the number of layers in \textbf{S-LSTM$_\text{c-gl}$} did not give any gain (see Table 2 in the Appendix).} From the results, we can make the following conclusions. First, {B-LSTM$_\text{rand}$} overfits extremely on the asynchronous datasets, giving the worst results among the LSTMs. Second, pretrained vectors help  to achieve better results, however, compared to the off-the-shelf vectors, our conversational word vectors yield much higher $F_1$, especially, in the asynchronous datasets that are smaller in size (5 - 11\% absolute gains). This demonstrates that pretrained word embeddings provide an effective method to perform semi-supervised learning, when they are learned from relevant datasets. %Finally, LSTM cells generally perform better than GRU cells. 

{The \textbf{last block} shows the results of our models. It is evident that both H-LSTM and H-LSTM-CRF outperform other models in all the datasets except QC3 where the difference is very small. %\footnote{There is no significant difference between H-LSTM and B-LSTM in accuracy also as can be seen in the Appendix.} 
They also give the best $F_1$ reported so far on MRDA, outperforming the B-LSTM models of \newcite{joty-hoque:2016} and S-LSTM model of \newcite{Khanpour-coling-16}. When we compare the two models, we notice that H-LSTM outperforms H-LSTM-CRF in all the datasets. A reason for this could be that the contextual dependency is already captured by the upper LSTM layer and the data may be too small for the CRF to offer anything more.}

%\vspace{-0.1em}
\subsection{Experiments on Domain Adaptation} \label{sec:eval-adv-adaptation}

%\vspace{-0.5em}
\paragraph{Settings.} {We compare our adversarial adaptation method with three baseline methods: Transfer, Merge and Fine-tune. \textbf{Transfer} models are trained on the source (MRDA) and tested on the target (QC3, TA, BC3). Our \textbf{adversarial unsupervised adaptation} method is comparable to the transfer method as they use labeled data only from the source domain. %\footnote{Note that our adversarial unsupervised adaptation method also uses unlabeled data from the target domain.} 
In \textbf{Merge}, models are trained on the concatenated training set of source and target datasets. \textbf{Fine-tune} is a widely used adaptation method for neural models \cite{Chenhui-coling18}. In this method, we first train a model on the source domain until convergence, then we fine-tune it on the target by training it further on the target dataset. Both merge and fine-tune are comparable to our \textbf{semi-supervised/supervised adaptation} as these methods use labeled data from the target domain. For {semi-supervised} experiments, we take smaller subsets (\eg\ 25\%, 50\%, and 75\% of the labeled data) from the target domain.} 

We also compare our method with Neural Structural Correspondence Learning or \textbf{Neural SCL} \cite{ziser-reichart:2017:CoNLL}, which is another domain adaption method in the neural framework. We used the implementation made available by the authors.\footnote{\href{https://github.com/yftah89/structural-correspondence-learning-SCL}{{https://github.com/yftah89/structural-correspondence-learning-SCL}}} For training our adaptation models, we use SGD (Algorithm 1 in the Appendix) with a momentum term of 0.9 and a dynamic learning rate as suggested by \newcite{Ganin:2016}.

%transfer and concatenation 
%Similarly, our \textbf{unsupervised adaptation} models use labeled data only from the source domain, thus they are comparable. 
%On the other hand, in concatenation models (\textbf{Concat}), we merge the source and target (labeled) training data to train the model. 

%So, this is directly comparable to our \textbf{supervised adaptation}. 

%We  which is adjusted during the training. For choosing the domain adaptation hyperparameter values, we followed the values used by \cite{Ganin:2016}. 

\vspace{-0.5em}
\paragraph{Results.}

\begin{table}[tb!]
\centering
\scalebox{0.64}{\begin{tabular}{l|l|c|c|c}
\toprule
%   & \multicolumn{4}{c}{\textbf{Adversarial Domain Adaptation}} \\
%\midrule
\textbf{Method}  & \textbf{Model} & \textbf{QC3}  & \textbf{TA} & \textbf{BC3} \\
\hline
\multirow{ 6}{*}{\textbf{Transfer}}  & \add{SVM} & $17.78_{\pm0.00}$  & $20.44_{\pm0.00}$  & $17.85_{\pm0.00}$ \\ 
& \add{FFN} & $46.91_{\pm0.00}$  & $56.30_{\pm0.00}$  & $46.74_{\pm0.00}$    \\ 
& \add{S-LSTM}  &  $49.89_{\pm1.29}$  & $62.52_{\pm1.49}$  & $36.36_{\pm1.28}$   \\ 
& {B-LSTM}  & $50.50_{\pm0.91}$  & $\mathbf{65.47_{\pm0.62}}$  & $35.92_{\pm0.62}$     \\ 
& \add{H-LSTM}  & $50.22_{\pm0.64}$  & $64.43_{\pm0.52}$  & $35.11_{\pm1.64}$     \\ 
& \add{H-LSTM-CRF} & $\mathbf{50.83_{\pm0.70}}$  & $63.80_{\pm0.81}$  & $34.45_{\pm1.42}$   \\ 
\hline
\multirow{ 5}{*}{\parbox{1.7cm}{\textbf{Unsup. adapt}}} & \add{Neural SCL} &$37.73_{\pm0.92}$ & $53.98_{\pm0.33}$ & $\mathbf{46.90_{\pm0.89}}$  \\
%\cline{2-5}
 & \add{Adv-S-LSTM} & $43.36_{\pm1.44}$  & $48.51_{\pm0.51}$  & $42.05_{\pm0.21}$ \\
 & \add{Adv-B-LSTM} & $47.39_{\pm0.74}$  & $58.49_{\pm1.29}$  & $32.86_{\pm1.35}$ \\
 & \add{Adv-H-LSTM} & $46.53_{\pm1.48}$  & $52.90_{\pm1.20}$  & $31.36_{\pm1.91}$ \\
& {Adv-H-LSTM-CRF} & $47.06_{\pm1.24}$  & $61.58_{\pm0.78}$  & $29.54_{\pm1.06}$  \\
\hline
\hline
\multirow{ 4}{*}{\parbox{1.6cm}{\textbf{Merge (50\%)}}} & \add{S-LSTM} &  $55.39_{\pm0.29}$  & $67.79_{\pm0.15}$  & $50.82_{\pm0.98}$\\
& {B-LSTM} & $55.08_{\pm0.67}$  & $68.99_{\pm0.35}$  & $51.05_{\pm0.60}$  \\
& \add{H-LSTM} & $51.74_{\pm0.44}$  & $69.09_{\pm0.64}$  & $47.82_{\pm1.47}$ \\ 
 & \add{H-LSTM-CRF} & $50.92_{\pm0.48}$  & $68.66_{\pm0.12}$  & $48.58_{\pm0.59}$  \\ 
\hline
\multirow{ 4}{*}{\parbox{1.6cm}{\textbf{Fine-tune (50\%)}}} 
& {S-LSTM} & $53.94_{\pm1.04}$  & $66.07_{\pm0.67}$  & $51.73_{\pm1.53}$ \\
& {B-LSTM} & $54.81_{\pm0.48}$  & $68.43_{\pm0.51}$  & $52.26_{\pm0.82}$ \\
& {H-LSTM} & $54.34_{\pm0.71}$  & $69.16_{\pm0.66}$  & $50.81_{\pm0.97}$ \\ 
 & {H-LSTM-CRF} & $54.97_{\pm0.87}$  & $69.91_{\pm0.53}$  & $51.42_{\pm1.24}$  \\ 
\hline
\multirow{ 5}{*}{\parbox{1.6cm}{\textbf{Semisup. adapt (50\%)}}} & \add{{Neural SCL}} & $41.46_{\pm0.75}$ & $58.85_{\pm0.27}$ & $48.32_{\pm0.19}$\\
 & \add{{Adv-S-LSTM}} & $60.20_{\pm0.32}$  & $68.71_{\pm0.75}$  & $57.97_{\pm0.39}$\\
 & \add{{Adv-B-LSTM}} & $58.57_{\pm0.80}$  & $66.51_{\pm0.81}$  & $54.39_{\pm1.28}$ \\
 & \add{{Adv-H-LSTM}} & $60.19_{\pm1.10}$  & $69.43_{\pm0.64}$  & $58.39_{\pm1.12}$ \\
%& \add{{H-LSTM}} & {Semi-sup adapt ($\frac{1}{4}$)} & $58.44_{\pm2.91}$  & $65.57_{\pm0.74}$  & $56.30_{\pm2.91}$     \\
%& {Semi-sup adapt ($\frac{3}{4}$)} & $62.96_{\pm1.27}$  & $71.10_{\pm0.57}$  & $59.13_{\pm1.72}$ \\
%\midrule
%& \add{{H-LSTM-CRF}} & $61.64_{\pm1.03}$  & $68.83_{\pm0.40}$  & $57.69_{\pm1.22}$    \\
& \add{{Adv-H-LSTM-CRF}} & $\mathbf{61.81_{\pm0.63}}$  & $\mathbf{70.34_{\pm0.62}}$  & $\mathbf{59.43_{\pm1.41}}$ \\
%& {Semi-sup adapt ($\frac{3}{4}$)} & $61.99_{\pm0.94}$  & $71.53_{\pm0.30}$  & $58.91_{\pm1.34}$ \\
\hline
\hline
\multirow{ 4}{*}{\parbox{1.6cm}{\textbf{Merge (100\%)}}} & \add{S-LSTM} & $59.18_{\pm1.40}$  & $66.93_{\pm1.70}$  & $54.87_{\pm1.47}$ \\ 
& {B-LSTM} & $58.33_{\pm0.84}$  & $70.12_{\pm0.39}$  & $55.89_{\pm0.89}$ \\ 
& \add{H-LSTM} & $59.85_{\pm0.57}$  & $70.40_{\pm0.41}$  & $57.19_{\pm0.87}$ \\ 
 & \add{H-LSTM-CRF} & $59.53_{\pm0.66}$  & $69.88_{\pm0.68}$  & $56.04_{\pm1.15}$ \\ 
\hline
\multirow{ 4}{*}{\parbox{1.6cm}{\textbf{Fine-tune (100\%)}}} 
& {S-LSTM} & $56.67_{\pm0.85}$  & $67.41_{\pm0.34}$  & $56.40_{\pm0.44}$  \\ 
& {B-LSTM} & $59.74_{\pm0.53}$  & $69.87_{\pm0.82}$  & $57.09_{\pm1.14}$ \\ 
& {H-LSTM} &  $60.12_{\pm0.44}$  & $70.96_{\pm0.61}$  & $58.09_{\pm1.03}$ \\ 
 & {H-LSTM-CRF} &  $59.95_{\pm0.59}$  & $70.44_{\pm0.76}$  & $57.17_{\pm0.97}$ \\ 
\hline
\multirow{ 5}{*}{\parbox{1.6cm}{\textbf{Sup. adapt}}} & \add{{Neural SCL}} & $43.35_{\pm0.30}$ & $60.40_{\pm0.23}$ & $48.88_{\pm0.70}$ \\
 & \add{{Adv-S-LSTM}} & ${61.15_{\pm0.48}}$  & $70.33_{\pm0.66}$  & ${59.19_{\pm0.67}}$\\
 & \add{{Adv-B-LSTM}} & ${60.60_{\pm0.68}}$  & $69.30_{\pm0.50}$  & ${60.12_{\pm1.26}}$ \\
 & \add{{Adv-H-LSTM}} & $\mathbf{63.10_{\pm0.83}}$  & $72.82_{\pm0.59}$  & $\mathbf{60.38_{\pm1.07}}$ \\
& \add{{Adv-H-LSTM-CRF}} & $62.24_{\pm0.74}$  & $\mathbf{73.04_{\pm0.38}}$  & $59.84_{\pm0.75}$ \\
\bottomrule
\end{tabular}}
\vspace{-0.3em}
\caption{\label{tab:results_async_adp} Domain adaptation results on our datasets. All models use \textbf{conversational word embeddings}. Results are averaged over (\textbf{2 folds$\times$5) 10 runs}.} %Transfer models are trained on source and tested on target. Unsupervised adaptation models use unlabeled data from the target domain.} 
%\vspace{-1.5em}
\end{table}

%\subsubsection{}
The adaptation results are shown in Table \ref{tab:results_async_adp}. We observe that without any labeled data from the target (\textbf{Unsup. adap}), our adversarial adapted models (Adv-H-LSTM, Adv-H-LSTM-CRF) perform worse than the transfer baseline in all three datasets. In this case, since the out-of-domain labeled dataset (MRDA) is much larger, it overwhelms the model inducing features that are not relevant for the task in the target domain. However, when we provide the models with some labeled in-domain examples in the \textbf{semi-supervised (\textbf{50\%})} setting, we observe about 11\% absolute gains in QC3 and BC3 over the corresponding Merge baselines, and {7 - 8\%} gains over the corresponding Fine-tune baselines. As we add more target labels (\textbf{100\%}), performance of our adapted models (\textbf{Sup. adap}) improve further, yielding sizable improvements {($\sim$ 3\% absolute)} over the corresponding baselines in all datasets. Also notice that our adversarial adaptation  outperforms Merge and Fine-tune methods for all  models over all datasets, showing its effectiveness.

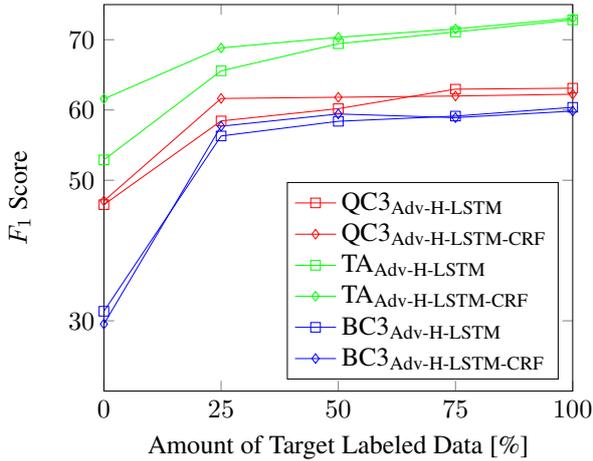
\begin{figure}[t!]
\centering
\begin{tikzpicture}[scale=0.9]
\begin{axis}[
    xlabel={Amount of Target Labeled Data [\%]},
    ylabel={$F_1$ Score},
    xmin=0, xmax=100,
    ymin=20, ymax=75,
    xtick={0,25,50,75,100},
    ytick={30,50,60,70,100},
    legend pos=south east,
    legend cell align={left},
    ymajorgrids=false,
    grid style=dashed,
]
 
\addplot[
    color=red,
    mark=square,
    ]
    coordinates {
    (0, 46.53)(25,58.44)(50,60.19)(75,62.96)(100,63.10) 
    };
\addplot[
    color=red,
    mark=diamond,
    ]
    coordinates {
    (0, 47.06)(25,61.64)(50,61.81)(75,61.99)(100,62.24) 
    };

\addplot[
    color=green,
    mark=square,
    ]
    coordinates {
    (0, 52.90)(25,65.57)(50, 69.43)(75,71.10)(100,72.82) 
    };
\addplot[
    color=green,
    mark=diamond,
    ]
    coordinates {
    (0, 61.58)(25,68.830)(50, 70.34)(75,71.53)(100,73.04) 
    };

\addplot[
    color=blue,
    mark=square,
    ]
    coordinates {
    (0,31.36)(25,56.30)(50,58.39)(75,59.13)(100,60.38) 
    };
\addplot[
    color=blue,
    mark=diamond,
    ]
    coordinates {
    (0,29.54)(25,57.69)(50,59.43)(75,58.91)(100,59.84) 
    };
\legend{QC3$_\text{Adv-H-LSTM}$, QC3$_\text{Adv-H-LSTM-CRF}$, TA$_\text{Adv-H-LSTM}$,TA$_\text{Adv-H-LSTM-CRF}$,BC3$_\text{Adv-H-LSTM}$,BC3$_\text{Adv-H-LSTM-CRF}$}
\end{axis}
\end{tikzpicture}
\vspace{-1em}
\caption{{$F_1$ with varying amount of target labels.}}
\label{fig:data-vs-f1}
\vspace{-1em}
\end{figure}

Figure \ref{fig:data-vs-f1} presents the $F_1$ scores of our  adapted models with varying amount of labeled data in the target domain. We notice that the largest improvements for all three datasets come from the first 25\% of the target labels. The gains from the second quartile are also relatively higher than the last two quartiles for TA and BC3. Another interesting observation is that H-LSTM-CRF performs better in unsupervised and semi-supervised settings (\ie\  with less target labels). In other words, H-LSTM-CRF adapts better than H-LSTM with small target datasets by exploiting the tag dependencies in the source. As we include more  labeled data from the target, H-LSTM catches up with H-LSTM-CRF. Surprisingly, Neural SCL performs the worst. We suspect this is due to the mismatches between pivot features of the source and target domains.

%As we include more target data, the H-LSTM with Softmax catches up with H-LSTM-CRF. This tells us that the H-LSTM-CRF adapts better than H-LSTM for small target datasets by exploiting the output (label) dependencies in the source (MRDA) domain.  

%\subsubsection{Results and discussion on CRF}

%\blue{CRF is well-known for capturing inter-sentence dependencies in a conversation. In the domain adaptation training, hierarchical LSTM model  However, in supervised settings, the hierarchical LSTM model without CRF  and performs better in QC3 and BC3 datasets.} 

%(In these cases, because there is more data available, the hierarchical LSTM without CRF learns the sentence dependencies better and outperforms the model with CRF.}

%   This demonstrates the effectiveness of our adversarial domain adaptation method. In the future, it will be interesting to compare adversarial training with other domain adaptation methods.

%of our adapted models

%The reason is that the source labeled dataset (\mrda) dominate the model by inducing the features that are not related to the task in the target domain. 

%   (last row),  gains merge model (second row) in all three datasets. Remarkably, the absolute improvements in $F_1$ for BC3 and TA are more than 11\% and 3\%, respectively.

%On the other hand, when we give more labeled target data to the model in semi-supervised and supervised adaptation settings, $F_1$ scores get the improvement of 10.07\% in QC3, 1.68\% in TA, 9.85\% in BC3, and 3.25\% in QC3, 2.64\% in TA, 3.19\% in BC3 respectively compared to their non adapted baseline models. 

{If we compare our \textbf{adaptation results} with the \textbf{in-domain results} in Table \ref{tab:indom-results}, we notice that using the same amount of labeled data in the target, our supervised adaptation gives 3-4\% gains across the datasets. Our semi-supervised adaptation using half of the target labels (50\%) also outperforms the in-domain models that use all the target labels.}    

%\red{Now, if we compare our supervised adaptation results in Table \ref{tab:results_async_adp} with the in-domain results in Table \ref{tab:indom-results}, we see our adaptation method gives about 3 - 4\% absolute gains in all three datasets, whereas simple concatenation gives hardly any gain.} 

\begin{figure}[tb!]
\vspace{-2em}
% \scalebox{1.0}{
%\hspace{-1.3em}
\begin{subfigure}[b]{1.1\linewidth}
\centering
\includegraphics[width=1\linewidth]{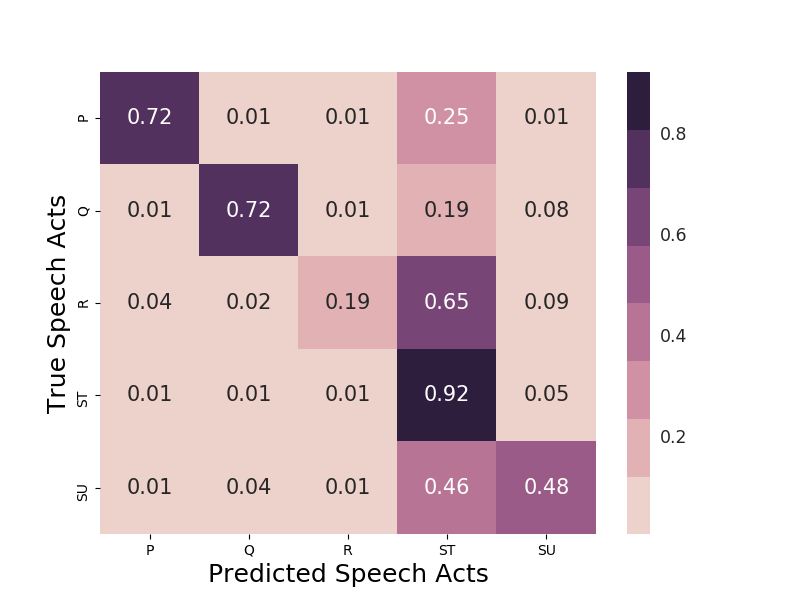}
%    \vspace{-0.2em}
  \caption{Non-adapted H-LSTM}
  \label{fig:conf-MLP}
\end{subfigure}
\begin{subfigure}[b]{1.0\linewidth}
\vspace{-0.2em}
\centering
\includegraphics[width=1.1\linewidth]{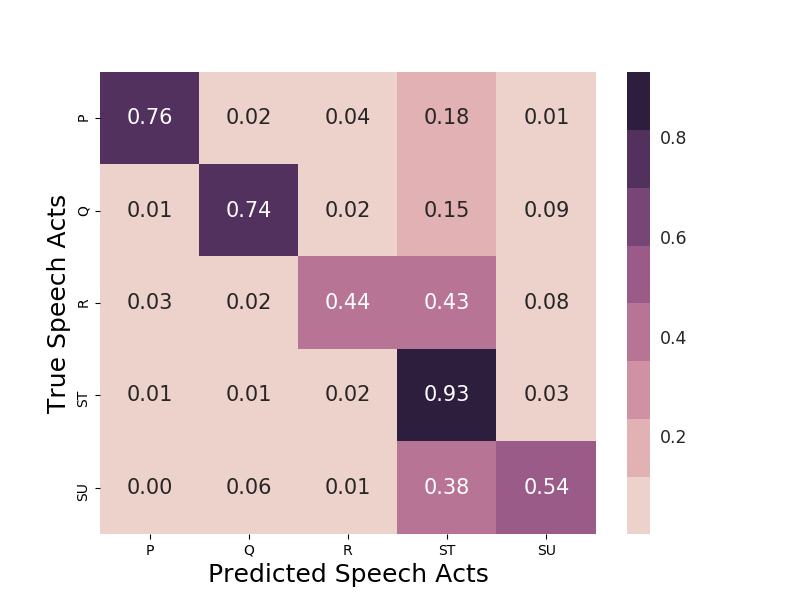}
%    \vspace{-0.2em}
  \caption{\add{Adapted H-LSTM}}
  \label{fig:conf-BLSTM}
\end{subfigure}
% }
%\vspace{-0.8em}
\caption{\small{Confusion matrices on the combined test sets.}}% for (a) Non adapted H-LSTM and (b) {Adapted H-LSTM} P stands for \pol, Q for \ques, R for \res, ST for \st, and SU stands for \sug.}}
\label{fig:conf}
\vspace{-1.2em}
\end{figure}

{To further analyze the cases where our adapted models make a difference, {Figure \ref{fig:conf}} shows the confusion matrices for the adapted H-LSTM and the non-adapted H-LSTM on the concatenated testsets of QC3, TA, and BC3. In general, our classifiers get confused between \emph{Response} and \emph{Statement}, and between \emph{\sug\ } and \emph{Statement} the most. We noticed similar phenomena in the human annotations, where annotators had difficulties with these three acts. It is however noticeable that the adapted H-LSTM is less affected by class imbalance, and it can detect the \emph{\sug} and \emph{\pol} acts more correctly than the non-adapted one.}

\begin{comment}
\begin{table}[tb!]
\centering
\scalebox{0.88}{\begin{tabular}{l|l|c|c|c}
\toprule
   & Training Regime & QC3  & {TA } & {BC3} \\
\midrule
%   EasyAdapt \cite{Daume-2007}\\
%   EasyAdapt++ \cite{Daume-2010}\\
%   SCL \cite{Blitzer:2006} & \\
   Neural SCL & \\
\midrule
Unsup adap   & \add{{H-LSTM}} &  \\
Semisup adap & \add{{H-LSTM}} &   \\
\bottomrule
\end{tabular}}
\vspace{-0.5 em}
\caption{\label{tab:results_async_adp_comp} Comparison with other domain adaptation methods.} 
\end{table}
\end{comment}

%% commented out for draft submission.. revise it..

%\vspace{-0.4em}
\section{Conclusion}
\label{sec:conclusion}
%\vspace{-0.4em}
We proposed an adaptation framework for speech act recognition in asynchronous conversation. Our base model is a hierarchical LSTM encoder with a Softmax or CRF output layer, which achieves state-of-the-art results for in-domain training. Crucial to its performance is the conversational word embeddings. We adapted our base model with adversarial training to effectively leverage out-of-domain meeting data, and to improve the results further. A comparison with existing methods and baselines in different training scenarios demonstrates the effectiveness of our approach. 

%\red{ demonstrates the effectiveness of our approach. include settings, ranging from unsup to sup.} 

%In future, we would like to explore ways to improve the conversational word embeddings using contexts like ELMo \cite{Peters:2018}.  

%---------------------------------------------- A c k n o w l e d g m e n t s
\section*{Acknowledgments}

Shafiq Joty would like to thank the funding support from MOE Tier-1 (Grant M4011897.020).

\bibliography{SA}
\bibliographystyle{acl_natbib}

%\appendix

\end{document}